\pdfoutput=1
\documentclass[11pt]{article}

\usepackage{acl}

\usepackage{times}
\usepackage{latexsym}
\usepackage[colorinlistoftodos,prependcaption,textsize=tiny]{todonotes}
\usepackage[normalem]{ulem}

\usepackage[T1]{fontenc}
\usepackage[utf8]{inputenc}
\usepackage{microtype}
\usepackage{multirow}

\usepackage{booktabs}

\usepackage{makecell}
\usepackage{subcaption}

\usepackage{etoolbox,siunitx}
\renewcommand{\bfseries}{\fontseries{b}\selectfont}
\newrobustcmd{\B}{\bfseries}

\usepackage{enumitem}
\usepackage{tablefootnote}
\usepackage{amssymb}
\usepackage{amsmath}
\usepackage[nameinlink]{cleveref}
\usepackage{mdframed,listings}
\def\avgwbleu{\textsc{Avg BLEU}}
\DeclareMathOperator{\Acc}{Acc}

\definecolor{darkgreen}{RGB}{0, 130, 0}

\title{Selectively Answering Visual Questions}

\author{
Julian Martin Eisenschlos$^{1,2}$,
Hernán Maina$^{2,3}$,
Guido Ivetta$^{2,3}$,
Luciana Benotti$^{2,3}$ \\
Google DeepMind$^1$ \\
Universidad Nacional de Córdoba $^2$,
CONICET, Argentina $^3$ \\
\texttt{\{julian.eisenschlos,hernan.maina,guidoivetta\}@mi.unc.edu.ar}\\
\texttt{\{luciana.benotti\}@unc.edu.ar}
}
\begin{document}
\maketitle
\begin{abstract}
Recently, large multi-modal models (LMMs) have emerged with the capacity to perform vision tasks such as captioning and visual question answering (VQA) with unprecedented accuracy.
Applications such as helping the blind or visually impaired have a critical need for precise answers.
It is specially important for models to be well \emph{calibrated} and be able to quantify their uncertainty in order to \emph{selectively} decide when to answer and when to abstain or ask for clarifications.
We perform the first in-depth analysis of calibration methods and metrics for VQA with in-context learning LMMs.
Studying VQA on two answerability benchmarks, we show that the likelihood score of visually grounded models is better calibrated than in their text-only counterparts for in-context learning, where sampling based methods are generally superior, but no clear winner arises.
We propose \avgwbleu, a calibration score combining the benefits of both sampling and likelihood methods across modalities. 
\end{abstract}

\section{Introduction}

Reliable Visual Question Answering (VQA) systems should provide a \emph{confidence estimate} of their own predictions.
This introspective skill allows users to decide when to trust or double check its outputs, or lets the system itself \emph{selectively} decide when to gather more information in order to provide accurate responses.

While this problem has been studied extensively for classification models~\cite{guo2017calibration} and more recently also for text generation~\cite{flexqa}, VQA systems have been under-explored.
The combination of multiple input modalities can introduce different sources of uncertainty due to incorrectly framed or focused images.
This is specially true when images are taken by people with no or limited vision, as is the case for the VizWiz-VQA dataset~\cite{gurari2018vizwiz}.
Furthermore the questions in VizWiz-VQA are spoken and therefore more conversational and contextual, which can introduce additional denotational uncertainty due to ambiguity when the question is under-specified because of shared common ground~\cite{flexqa}.
The multiple crowd-worker annotations in the dataset allow us also to identify confusing or unanswerable questions. It also allows us to see cases when even the unique answer to an unambiguous question can have multiple equivalent surface forms, or be expressed with different levels of specificity, for example in the month and year, or the full date for an expiration date.

% https://docs.google.com/drawings/d/1D7N7iZOsuxo9YJVXp_R2DYfwIfWTtlVEYqqCqU1J6cw/edit
\begin{figure}[t]
    \centering
    \includegraphics[width=1.0\linewidth]{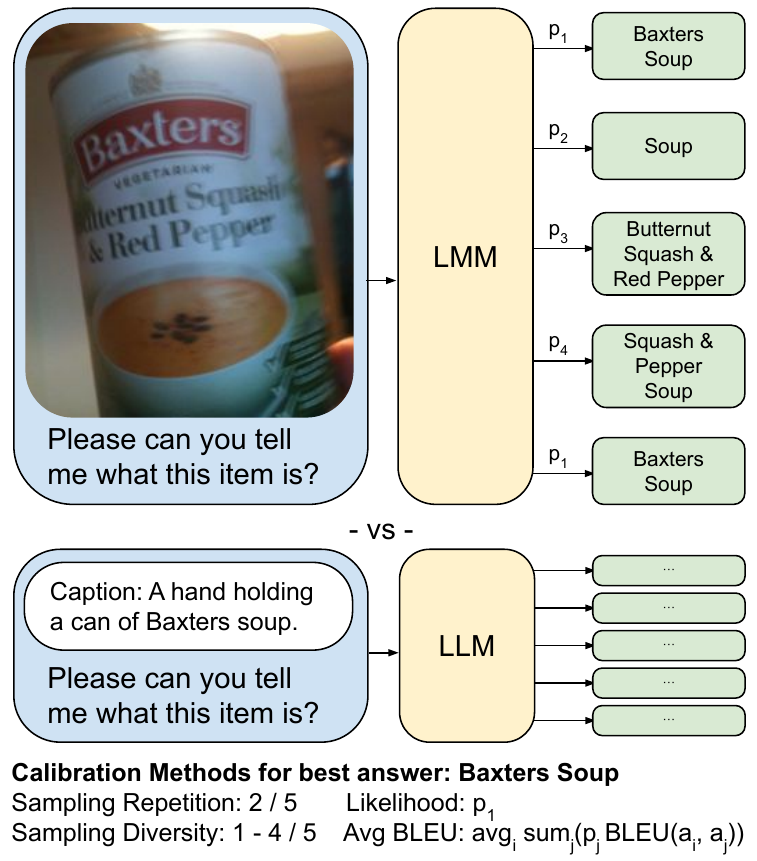}
    \vspace*{-8mm}
    \caption{Sampled outputs from an LMM on a VizWiz-VQA example~\cite{gurari2018vizwiz} are used to measure \emph{model calibration}.
    In this paper we contrast the LMM calibration results against LLMs that only sees the image caption.
    % Middle: Asking the same question but using only a caption of the image. Bottom: Comparison of different calibration methods. 
    Sampling based methods struggle to measure uncertainty, motivating our proposed \avgwbleu{} as a \emph{confidence estimate}.}
    \label{fig:diagram}

\end{figure}

\begin{figure}[t]
    \centering
    \includegraphics[width=1.0\linewidth]{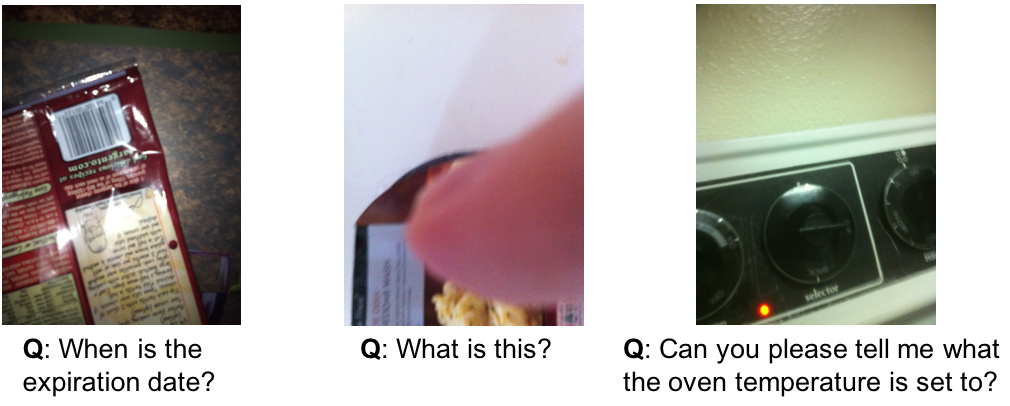}
\vspace*{-7mm}
\caption{Unanswerable questions in VizWiz-VQA. Many are not answerable due to low quality images. 
Sometimes the intent of the question poser, and therefore the correct answer, cannot be inferred. 
% \lb{How about we mix unanswerable and answerable? I think unsuitable images do not need to be illustrated, they complicate the story unnecesarily.}
}
    \label{fig:examples}
% \vspace*{-5mm}
\end{figure}

Our goal is to study which methods from the literature can help calibrate state-of-the-art VQA systems, and build a \emph{selective} VQA setup that can trust  its confidence estimates when answering.
We see this as a necessary step towards systems that can be used in real life scenarios.
\Cref{fig:diagram} introduces three of the methods used in~\citet{flexqa} as well as our proposed \avgwbleu \ (inspired in~\citet{wan2023universal}) \ on an example from the VizWiz dataset~\cite{gurari2018vizwiz}. %\avgwbleu \ aims to quantify the spread of possible model predictions.

Our contributions can be summarized as follows:

\noindent\textbf{(a)} Through the use of several metrics, we study how the calibration methods from text QA perform when applied in two VQA answerability benchmarks and find that likelihood based scoring is better for LMMs than LLMs, without a clear winner.

\noindent\textbf{(b)} We propose a new sampling based method that tackles limitations and improves all metrics significantly. Concretely, it improves coverage at $80\%$ accuracy by $5$ points for the best LMM and coverage at $70\%$ accuracy by $8$ points for the best LLM.

\section{Related Work}

Our paper studies VQA systems with unanswerable question through the VizWiz-VQA~\cite{gurari2018vizwiz} and UNK-VQA~\cite{guo2023unkvqa} datasets.
We show some examples in \Cref{fig:examples}.
Through the introduction of \emph{unanswerable} responses from the annotators, Gurari et al.~study a possible form of calibration as binary detection of such images.
This setting is however limited in that a model (or person) can assess a question as \emph{answerable} and be \emph{uncertain} about whether the answer is a good one, as observed by~\citet{Chen_2023_ICCV}.

On the topic of estimating uncertainty, several approaches appear in the literature. \citet{collier2021correlated} incorporate a latent trainable variable for the covariance matrix among classes. 
The use of additional binary classifiers--- so-called \emph{selectors}---was investigated for text QA~\cite{kamath-etal-2020-selective, desai-durrett-2020-calibration, jiang-etal-2021-how}.
This approach was extended to VQA by \citet{reliable-vqa} and \citet{dancette2023improving}. They focus on fine-tuned models and the use of selectors.
We argue that, with the increasing availability and adoption of LMMs, it is important to study calibration in zero and few-shot scenarios, which pose unique challenges for evaluation~\cite{maynez-etal-2023-benchmarking}.

\citet{flexqa} study different sampling-based methods to calibrate LLMs in a zero-shot fashion.
It focuses on entity based QA, where there is little differences in the surface form of equivalent answers.
Due to the open-ended problem, the answers in VizWiz-VQA and UNK-VQA can vary, as illustrated in \Cref{fig:diagram}. This variation needs to be considered when designing schemes to measure uncertainty when evaluating multiple samples.

\begin{table*}[t]
    \centering
    % \footnotesize
    \resizebox{1.0\textwidth}{!}{%
    \begin{tabular}{l|rrrr|rrrrr|rrrrr|rrrrr|}
 & \multicolumn{9}{c|}{\textbf{Large Multimodal Models (LMMs)}} & \multicolumn{10}{c|}{\textbf{Large Language Models (LLMs)}} \\
Model & \multicolumn{4}{c|}{\textbf{LLaVA 13B} (61\% ac@34\% trig)} & \multicolumn{5}{c|}{\textbf{Flamingo 3B} (20.5\% ac@71.7\% trig)} & \multicolumn{5}{c|}{\textbf{PaLM 2 Bison} (33.6\% ac@78.8\% trig)} & \multicolumn{5}{c|}{\textbf{Falcon} (18.5\% ac@87.0\% trig)} \\
Method & AUC & ECE$\downarrow$ & C@70 & C@80 & AUC & ECE$\downarrow$ & C@60 & C@70 & C@80 & AUC & ECE$\downarrow$ & C@60 & C@70 & C@80 & AUC & ECE$\downarrow$ & C@60 & C@70 & C@80 \\
\midrule
\avgwbleu & \textbf{70.5} & 28.6 & \textbf{73.8} & \textbf{33.3} & \underline{88.1} & 19.5 & \underline{18.1} & \underline{9.5} & \textbf{4.7} & \underline{73.6} & \textbf{9.8} & \textbf{26.4} & \textbf{18.4} & \textbf{11.2} & \underline{85.1} & \textbf{13.5} & \textbf{12.1} & \textbf{4.6} & \underline{1.1}\\
\midrule
Likelihood & 68.8 & \underline{7.9} & 71.6 & 20.8 & \textbf{88.7} & 17.3 & \textbf{22.1} & \textbf{12.1} & 0.5 & 57.5 & \underline{20.6} & 0.0 & 0.0 & 0.0 & 73.8 & 64.2 & 7.2 & \underline{1.7} & \textbf{1.4}\\
Diversity & \underline{70.4} & \textbf{6.8} & 67.2 & 8.7 & 77.6 & \underline{12.4} & 11.4 & 7.2 & 3.7 & \textbf{75.0} & 39.9 & 10.1 & \underline{10.1} & \underline{10.1} & \textbf{86.2} & \underline{31.3} & \underline{11.8} & 0.0 & 0.0\\
Repetitions & 69.3 & 9.9 & \underline{73.2} & \underline{28.4} & 76.7 & \textbf{5.4} & 7.9 & 7.9 & \underline{4.4} & 72.7 & 31.8 & \underline{21.6} & \underline{10.1} & \underline{10.1} & 84.0 & 34.0 & \underline{11.8} & 0.0 & 0.0 \\
\end{tabular}

    }
    \caption{Calibration metrics on VizWiz-QA comparing LMMs on the left and LLMs on the right (by using gold image captions).
    We use $4$-shots, except for LLaVa which only supports $0$-shot.
    Best and second best values are \textbf{bolded} and \underline{underlined} respectively. 
    Likelihood lags behind sampling methods by more than $10$ points for most metrics on LLMs but can surpass them in LMMs, although there is no clear winner.
    % \lb{Do you mean "Likelihood lags behind for LLMs but is comparable to \avgwbleu\ in LMMs?"}
    % Sampling repetition and diversity suffer from relying on exact match between answers.
    \avgwbleu{} combines both methods and performs above or comparable to the rest except for ECE, which can be fixed post-hoc by re-scaling.}
    \label{tab:results}
    \vspace*{-2mm}
\end{table*}

\begin{table}[t]
    \centering
    % \footnotesize
    \resizebox{0.48\textwidth}{!}{%
    \begin{tabular}{l|rrrr|rrrrr|}
Model & \multicolumn{4}{c|}{\textbf{LLaVA 13B} (20\% ac@65\% trig)} & \multicolumn{5}{c|}{\textbf{Flamingo 3B} (12.5\% ac@48.0\% trig)} \\
Method & AUC & ECE$\downarrow$ & C@30 & C@40 & AUC & ECE$\downarrow$ & C@20 & C@30 & C@40 \\
\midrule
\avgwbleu & \underline{61.0} & \textbf{27.3} & \textbf{26.4} & \textbf{1.3} & \underline{71.9} & \underline{9.9} & \textbf{50.3} & \underline{17.4} & \underline{2.4}\\
\midrule
Likelihood & 60.1 & \underline{49.5} & \underline{22.1} & \textbf{1.3} & \textbf{72.4} & \textbf{3.8} & \underline{48.6} & \textbf{18.4} & \textbf{5.2}\\
Diversity & 60.7 & 52.4 & 21.3 & 0.0 & 69.0 & 23.6 & 37.8 & 3.5 & 0.0\\
Repetitions & \textbf{61.4} & 51.3 & 21.3 & 0.0 & 69.1 & 24.8 & 32.3 & 3.5 & 0.0 \\
\end{tabular}
    }
    \caption{Calibration metrics on UNK-QA.
    Best and second best values are \textbf{bolded} and \underline{underlined} respectively. 
    As is the case of VizWiz-VQA, likelihood is comparable or better than sampling methods.
    \avgwbleu{} performs above or comparable to the alternatives.}
    \label{tab:unk_results}
\end{table}

\section{Methods}

A QA system is said to be well-calibrated when each prediction has a confidence score which can help assess how often it is correct.
Formally, the paradigm of \emph{selective QA} uses a scoring function $s$ that attaches to each QA pair $(q, a)$ a numeric score $s(q, a)$. The score can be compared with a threshold $\tau$ so that the system answers---aka \emph{triggers}---when $s(q, a) > \tau$ or abstains otherwise.
As shown in \Cref{fig:diagram}, various scoring methods can be used. We present first the methods from~\citet{flexqa} for text-only QA, and then our proposal.

\paragraph{Likelihood} The likelihood-based calibration uses the output language model score $p(a|q)$ computed using the chain rule: $p(a|q) = \prod_{i=1}^n p(t_i | t_1, \cdots, t_{i-1}, q)$ where the $t_i$ are the tokens that form the answer $a$. 

\paragraph{Sampling Repetition} 
Based on~\citet{wilcox-1973-iqv}'s \emph{Variation Ratio} we compute the frequency of the most sampled output divided by the total number of samples, which coincides with the probability of the mode of the empirical distribution. 
When more samples agree with each other, the answer can be considered to be more trustworthy.
% We count the number of samples which \textit{exactly} match the greedy answer. We experimented briefly with more
% sophisticated approaches based on the BERT based answer equivalence model from \citet{bulian-etal-2022-tomayto} as well as the summed F1 score across sampled answers and the greedy answers, but these approaches did not meaningfully improve the calibration.

\paragraph{Sampling Diversity}
%A second IQV considers the proportion of distinct elements among the samples.
% As above, we sample the model multiple times. However, instead of considering a single answer appearing multiple times as the measure of confidence, we use having fewer answers other than the greedy answer produced. 
%For instance, producing ``Joe Biden'' seven times, ``Donald Trump'' once, and ``Barack Obama'' twice would be more confident under sampling repetition than producing ``Joe Biden'' six times and ``Barack Obama'' four times. Under sampling diversity, it is reversed, where having three answers is an indicator that it is less confident. In particular, we produce a calibration score by taking $1 - \frac{\mathrm{num\_unique}}{\mathrm{num\_samples}}$. Similar to before, we sample the model 10 times, so if there are 10 unique answers (each sample produces a different answer), then the calibration score would be 0. Note that while this works relatively well in practice, it is heavily dependent on the number of samples, as we do not expect the number of unique answers for the model to produce to scale linearly with the number of samples. As above, the computation of this score relies on a normalized exact string match but could instead employ more sophisticated methods. 
% \je{What is the actual metric here: something like percentage distinct? It was interesting to see that this metric is not listed at all in \url{en.wikipedia.org/wiki/Qualitative_variation} and instead some versions where the log or square root of the denominator is applied. This is presumably since otherwise this will be very unstable as the sample size increases.}
Computed as $1-\frac{\mathrm{\#unique}}{\mathrm{\#samples}}$, it is inversely proportional to the number of distinct samples and is zero if all samples are different.
% Note that while this works relatively well in practice, it is heavily dependent on the number of samples, as we do not expect the number of unique answers to scale linearly with the number of samples.

\paragraph{\avgwbleu} We propose an approach to grading the similarity among model answers that relies on averaging a measure of semantic similarity.
Since there are multiple answers, we consider the average among all the possible pairs as an estimate of the diameter or dispersion of the full set. This is inspired by the measure proposed by~\citet{wan2023universal} for LLMs.
Instead of ROUGE~\cite{lin-2004-rouge}, which is not sensitive to character n-grams, we use BLEU~\cite{papineni-etal-2002-bleu} and compute the pairwise weighted average $\frac{1}{k}\sum_{i, j} p(a_i|q)\text{BLEU}(a_i, a_j)$ over the set of $k$ distinct predicted answers.
Other similarity metrics are studied in \Cref{app:avg_bem}.
We fix the distance between a proper answer and \emph{unanswerable} to~$0$.

%
% More complex model-based measures of answer equivalence could also be used in the future, such as the one proposed by~\citet{bulian-etal-2022-tomayto} or the cosine similarity between embeddings. 

\section{Experiments}

In our experiments, we use the validation split of VizWiz-VQA with 4k instances.
%
% In total, it contains $20,523$ training questions, $8,000$ for testing, and 4,319 for validation. 
Each question has up to $10$ crowd-worker answers.
Examples can be seen in \Cref{fig:examples}.
We consider a question answerable if at least one crowd-worker annotated it as such. This corresponds to $75\%$ of the questions.

We also include the validation set of UNK-VQA~\cite{guo2023unkvqa}, consisting of 1K examples, synthetically constructed by modifying the images or text from VQA v2~\cite{balanced_vqa_v2}.

On the modeling side, we chose state-of-the-art LMMs LLaVA~\cite{llava2023}, Flamingo~\cite{flamingo2022}, and BLIP-2~\cite{li2023blip2}.
To compare the calibration of LMMs with LLMs, we leverage the human written captions for VizWiz images provided by~\citet{vizwizcaps2020}.
We chose to use gold captions to control the additional errors and uncertainty which could arise from model written ones.
Nevertheless we run a study in ~\Cref{app:model_captions} using captions from PaLI-X~\cite{chen2023palix} and found that similar results hold.
We evaluate state-of-the-art LLMs PaLM-2~\cite{anil2023palm} and Falcon~\cite{almazrouei2023falcon}.
We sample ten responses with a temperature of $0.7$. We consider the model to \emph{trigger} if the most likely answer (greedy) does not contain the sequence ``unanswerable''. 
The full prompts and BLIP-2 results can be seen in ~\Cref{app:exp}.
% Under this setup, the model triggers $30\%$ of the time with $69\%$ accuracy.
%
To simplify the analysis while staying consistent with the official metric defined in~\citet{VQA}, we consider a model answer correct if it matches at least one (instead of three) of the gold answers.

% \subsection{Metrics}
% % Selective question answering systems are evaluated on multiple criteria.
% There are many metrics for evaluating selective question answering systems, but fundamentally we want systems that (1) frequently return correct answers, and (2) rarely return incorrect answers. 
% These criteria are in conflict because any system can achieve zero error rate by always abstaining from answering.
% When the confidence score is a probability, we can also ask whether this probability is \emph{well-calibrated}, in the sense that a confidence score of $s(q,a)=\alpha$ implies $\alpha$ is the probability of $a$ being the correct answer to $q$.

% We break down our evaluations into two categories: calibration and selective prediction \cite{tran2022plex}. Calibration metrics roughly seek to evaluate the faithfulness of a system's confidence estimates, measuring the distance between the predicted confidence and the empirically observed probability of a correct answer. In contrast, selective prediction metrics establish a threshold for confidence scores, where predictions with confidence falling below this threshold are withheld. By varying this threshold, we are able to establish some desired trade off between maximizing coverage and accuracy.

We evaluate the different scoring methods defined in the previous section with intrinsic and extrinsic metrics that we introduce below, over the set of examples where each model triggers a response.

\paragraph{Expected Calibration Error (ECE)} Predictions are bucketed into ten same-sized bins, ranked by the confidence. We compute the mean absolute value of the distance between the average confidence score and the accuracy of predictions in each bin, averaged across all non empty bins.
This intrinsic evaluation interprets a confidence score to represent a probability, so it computes the difference in the predicted probability of being correct from the observed probability of being correct. ECE is therefore noisy and sensitive to re-scaling.

% \paragraph{Brier Score} This score is computed as the mean squared error between each examples confidence estimate and its binarized correctness value \cite{gneiting2007brier}.

\paragraph{ROC-AUC} Area under the receiver operating characteristic curve measures the ability as a binary classifier for correct and incorrect predictions by integrating over the curve of the rates of true and false positives. 
It can be interpreted as the probability that the score of a correct output chosen at random is higher than that of an incorrect output.

\paragraph{Coverage@Acc} While ECE and ROC-AUC assess absolute and relative calibration respectively, we want a metric closely aligned with selective QA.
Therefore, we compute the triggering rate the model can achieve if forced to maintain a certain accuracy among the responses where it triggers.
Formally, $\text{C@}\Acc$ is the maximum coverage such that the accuracy on the C\% of most-confident predictions is at least $\Acc\%$. For example, if C@$70 = 30$,  then the system is $70\%$ or more accurate on the top 30\% most-confident predictions. 
% To improve the discriminative power of the metric, 
We take $100$ coverage to mean not all of the examples, but where the model produced an answer.
% \lb{Is this example correct? If I look at table 1 I find it strange to say that Llava 13B is $60\%$ or more accurate on the top 100\% most-confident predictions.} \je{It is correct, with the caveat that hear we are looking at the predictions that trigger, so the 34\% where the model is not abstaining already, we could multiply everything by that number or explain it better. I will try to do the latter.}

% \begin{figure}[t]
%     \centering
%     \includegraphics[width=1.0\linewidth]{figures/calibration.pdf}
%     \caption{Calibration curves for the different scoring methods, according to the buckets in the computation of the Expected Calibration Error. Methods closer to the diagonal are better calibrated. Average BLEU achieves the best results.}
%     \label{fig:calibration}
% % \vspace*{-5mm}
% \end{figure}

We show the overall results for VizWiz in \Cref{tab:results}, including the baseline triggering rate and accuracy before incorporating calibration signals.
The strength of the models is validated by comparing to fine-tuned models on the task, where a Flan-T5 (Base)~\cite{chung2022scaling} model trained on all the available captions is $74.2\%$ accurate when it triggers, and a PaLI3-5~\cite{chen2023pali3} is $66.3\%$ accurate when it triggers. Details about the fine-tuning experiments are explained in ~\Cref{app:finetuning}.
% as shown in \Cref{tab:fine_tuned}.
%
For LLMs, the findings of~\citet{flexqa} hold, and likelihood is the worst calibration metric, repetition and diversity are superior to it. 
But, surprisingly, the picture changes for LMMs where we can see that repetition and diversity methods are not reliable.
In contrast, \avgwbleu{} works well across the board, matching or improving the best metric in most cases.
The same holds for UNK-VQA in \Cref{tab:unk_results}. 
We omit C@Acc for accuracy below the model overall accuracy since it is $100$.

We also evaluated model-based measures of semantic similarity in \Cref{app:avg_bem} beyond BLEU, 
but interestingly, we found that BLEU has strong performance while requiring negligible compute.

%
% The exception is for ECE, where we see that likelihood performs better, which is somewhat expected given that this requires a specific probabilistic interpretation of the score, which does not exist for BLEU.
%
% That said, the ECE metric can always be improved by introducing a calibration step that transforms the scores monotonically based on a control set.
%
% The full calibration curves can be seen in \Cref{fig:calibration}.
%
% We see that while the likelihood function is a strong baseline, the function is not monotonic, unlike the average pairwise BLEU.

%however we saw similar trends for other thresholds as well.

%\emph{Accuracy @ C\% Coverage} is computed as the QA system's exact-match accuracy on the subset of the $C\%$ of most confident predictions according to our evaluated system. Conversely, \emph{Coverage @ A\% Accuracy} is maximum $C$ such that the QA system's accuracy on the $C\%$ of most confident predictions achieves at least $A\%$ accuracy.

% \paragraph{Acc@C\%Cov and Cov@A\%Acc} These is the primary selective prediction metrics we consider in this work and are designed to mimic real world use cases of uncertainty estimates. \emph{Accuracy @ C\% Coverage} is computed as the QA system's exact-match accuracy on the subset of the $C\%$ of most confident predictions according to our evaluated system. Conversely, \emph{Coverage @ A\% Accuracy} is maximum $C$ such that the QA system's accuracy on the $C\%$ of most confident predictions achieves at least $A\%$ accuracy.

\subsection{Human vs. model written captions}\label{app:model_captions}

In order to restrict the noise and sources of uncertainty when evaluating LLMs for the VQA task, we opted to use gold captions.
However, it can be argued that a more realistic setup would require the use of automated captions.
With that goal, we evaluated the two LLMs on  captions generated by a fine-tuned PaLI-X model~\cite{chen2023palix}, which reported state-of-the-art results for the task when using an additional OCR module as input. Interestingly we observe improved results when using the model written captions (higher accuracy and triggering). Upon inspection of examples we see that the generated captions tend to be shorter and more concise, focusing on salient elements of the photo, which leads on average to better downstream answers for VizWiz where many questions are asking about salient objects (ex: \emph{``What is this?''}).

\begin{table}[!ht]
    \centering
    \footnotesize
    \resizebox{0.48\textwidth}{!}{%
    \begin{tabular}{l|rrrr|rrrr|}
Model & \multicolumn{4}{c|}{\textbf{PaLM 2 Bison} (45\% ac@79\% trig)} & \multicolumn{4}{c|}{\textbf{Falcon} (32\% ac@96\% trig)} \\
Method & AUC & ECE$\downarrow$ & C@60 & C@70 & AUC & ECE$\downarrow$ & C@60 & C@70 \\
\midrule
\avgwbleu & \textbf{75.4} & \underline{12.1} & \textbf{58.3} & \textbf{37.8} & \underline{74.8} & \textbf{7.9} & 22.1 & \textbf{1.5}\\
Likelihood & 63.1 & \textbf{11.0} & 28.5 & 4.1 & 67.9 & \underline{21.1} & 2.4 & \textbf{1.5}\\
Diversity & \underline{72.4} & 34.3 & \underline{42.6} & \underline{15.6} & \textbf{75.3} & 33.1 & \textbf{26.8} & 0.0\\
Repetitions & 67.2 & 27.0 & 37.9 & \underline{15.6} & 74.7 & 34.9 & \textbf{26.8} & 0.0 \\
\end{tabular}
    }
    \caption{Results on VizWiz-VQA using automated captions written by PaLI-X. Sampling based methods perform well except for ECE. \avgwbleu{} performs well across the board.}
    \label{tab:autocaptions}
\end{table}

As is the case for the manual captions, sampling based methods generally outperform likelihood by a large margin. 
The exception is on the ECE metric, which is expected given that it relies on the alignment between the scores and probabilities, can be fixed post-hoc, and is arguably the least practical one.
We also observe that, \avgwbleu{} is able to be better or comparable to best result in every case.

\subsection{Replacing BLEU for dense similarity}\label{app:avg_bem}

While BLEU has the advantage of fast execution, it can fail to capture more nuanced forms of similarities among answers.
We benchmarked two additional answer similarity metrics, BEM~\cite{bulian-etal-2022-tomayto} and BLEURT~\cite{sellam-etal-2020-bleurt}. 
Perhaps surprisingly, the results over the various tasks show very little effect, as seen in \Cref{tab:avg_bem}. 
We speculate that the variability among 10 crowd-worker answers diminishes the possible improvements, but it is possible that other VQA tasks could benefit from this approach making the additional compute cost worth the improved calibration.

\begin{table}[ht]
    \centering
    \footnotesize
    % \resizebox{1.0\textwidth}{!}{%
    \begin{tabular}{lrrr}
Method & AUC & ECE & C@60 \\
\hline
\hline
\textsc{Avg BLEU} & 88.1 & 19.5 & 18.1 \\
\textsc{Avg BEM} & 87.3 & 33.6 & 17.6 \\
\textsc{Avg BLEURT} & 88.6 & 18.8 & 22.3 \\
\end{tabular}
    % }
    \caption{Comparison of calibration metrics for Flamingo predictions when replacing BLEU for trained similarity metrics. Both BLEURT and BEM give marginal gains for coverage at the cost of increased latency.}
    \label{tab:avg_bem}
\end{table}

\begin{table}[t]
    \centering
    \footnotesize
    \resizebox{0.45\textwidth}{!}{%
    % \begin{tabular}{l|rrrr|rrrr|}
%  & \multicolumn{4}{c|}{LMM (39.3\% acc @ 11.3\% trig)} & \multicolumn{4}{c|}{LLM (24.6\% acc @ 10.8\% trig)} \\
% Model & \multicolumn{4}{c|}{\textbf{Blip2 Flan T5-XL}} & \multicolumn{4}{c|}{\textbf{Alpaca 3B}} \\
% Method & AUC & ECE$\downarrow$ & C@50 & C@60 & AUC & ECE$\downarrow$ & C@50 & C@60 \\
% \midrule
% \avgwbleu{} & 72.6 & \textbf{22.9} & \textbf{65.6} & 41.0 & 75.8 & \textbf{11.9} & \textbf{27.7} & \underline{15.4}\\
% Likelihood & 66.0 & 51.2 & 55.7 & 26.2 & 71.8 & 21.0 & \textbf{27.7} & \textbf{23.1}\\
% Diversity & \textbf{78.7} & \underline{29.5} & \textbf{65.6} & \textbf{49.2} & \underline{76.8} & 17.8 & 16.9 & 0.0\\
% Repetitions & \underline{75.8} & 32.6 & 57.4 & \underline{45.9} & \textbf{77.4} & \underline{16.6} & 15.4 & 0.0 \\
% \end{tabular}

\begin{tabular}{l|rrrrr|}
 & \multicolumn{5}{c|}{\textbf{Large Multimodal Model (LMM)}} \\
Model & \multicolumn{5}{c|}{\textbf{Blip2 Flan T5-XL} (39.3\% acc @ 11.3\% trig)} \\
Method & AUC & ECE$\downarrow$ & C@50 & C@60 & C@70 \\
\midrule
\avgwbleu{} & 72.6 & \textbf{22.9} & \textbf{65.6} & 41.0 & \textbf{23.0}\\
Likelihood & 66.0 & 51.2 & 55.7 & 26.2 & \textbf{23.0}\\
Diversity & \textbf{78.7} & \underline{29.5} & \textbf{65.6} & \textbf{49.2} & 0.0\\
Repetitions & \underline{75.8} & 32.6 & 57.4 & \underline{45.9} & 0.0 \\
\end{tabular}
    }
    \caption{Calibration metrics on VizWiz-QA comparing for BLIP-2. As in the main results for other models, we see \avgwbleu{} performs above or comparable to the alternatives for most metrics and well above likelihood.}
    \label{tab:more_results}
\end{table}

\section{Analysis and Discussion}\label{sec:analysis}

\paragraph{How can we enhance the reliability of a VQA system?} Based on the results presented in this work and previous literature in the effect of calibration scores in user facing applications~\cite{zhang-etal-2020-effect} we can make the following recommendations when implementing a VQA system:

\paragraph{(a)} Add a confidence score to the responses based on \avgwbleu{} that lets the users of VQA systems decide whether they can trust the system response or whether the question at hand is critical and needs a higher confidence in which case they can ask another person.
\paragraph{(b)} Automated captioning and LLMs on those captions can be useful to answer visual questions but likelihood is not a good calibration score in that case~\cite{cole2023selectively}.
\paragraph{Why are LMMs better calibrated?} We speculate that this has to do with the inability for diversity and repetition metrics to capture similarity beyond text exact match and the grounding effect of the multi-modality acting as a regularizer and distributing the likelihood of possible answers in a more meaningful way. We leave further experiments in this direction to future work.

The language modality---e.g. gold captions---is a human generated signal with high information density and communicative intent~\cite{rambow-walker-1994-role}. 
Other non-communicative natural signal modalities have heavy redundancy, noise, and low information density, as observed in other multi-modal tasks~\cite{wei-etal-2023-tackling}. 
We can speculate that this produces a regularization effect---meaning the model training objective forces it to spread its bets among many answers. 
Grounding~\cite{harnad:1990} the question to sections of the image becomes particularly hard for images taken by visually impaired people in VizWizVQA due to occlusions or lack of framing, introducing an additional source of uncertainty~\cite{ChenAG22}.

% Sighted people frame images centering the object of interest and avoiding occlusions.
% This is not possible for visually impaired people~\cite{ChenAG22}, introducing more uncertainty than traditional VQA and potentially strengthening the regularizer effect of  grounding~\cite{harnad:1990} more than previously observed for other multi-modal tasks~\cite{wei-etal-2023-tackling}.

% In this section we first explain how the model's answers are classified as correct or incorrect and then we illustrate the strengths are risks of LLMs and LLMs for our task. 

\paragraph{Can we measure accuracy better? }\label{sec:accuracy_methods}

In the previous experiments, we perform exact-match (EM) criteria to classify whether the generated answer was correct or not, following the standard VQA accuracy evaluation~\cite{VQA}. 

This approach led to instances where the model, when presented with a question such as ``What is my computer screen showing?'' produced the response ``A system restore''. Despite the relevance of the generated answer, it was erroneously classified as incorrect due to the stringent acceptance criteria that required matching exactly the annotated answers, such as ``system restore'', ``system restore message'', ``system restore pop up''.

\begin{table}[t]
    \centering
    \scriptsize
    % \resizebox{1.0\textwidth}{!}{%
    \begin{tabular}{lrrrr}
Method & EM & BLEU & Cos Sim & BEM~(\citeauthor{bulian-etal-2022-tomayto}) \\
\hline
Accuracy & 79\% & 81\% & 87\% & 87\% \\
\hline
\avgwbleu{} & \textbf{71} & \textbf{69} & \underline{73} & \underline{71} \\
Likelihood & 69 & 66 & 68 & 66 \\
Diversity & \underline{70} & \underline{67} & \textbf{75} & \textbf{72} \\
Repetitions & 69 & 66 & 70 & 69 \\
\end{tabular}

% \begin{tabular}{|l|rrrr|rrrr|}
% \hline
% Model       & \multicolumn{4}{c|}{LLaVa 13B}                                                                                & \multicolumn{4}{c|}{PaLM 2 Bison}                                                                             \\ \hline
% Method      & \multicolumn{1}{r|}{Exact Match} & \multicolumn{1}{r|}{BLEU} & \multicolumn{1}{r|}{Cos Similarity} & Bemscore & \multicolumn{1}{r|}{Exact Match} & \multicolumn{1}{r|}{BLEU} & \multicolumn{1}{r|}{Cos Similarity} & Bemscore \\ \hline
% Avg wBLEU   & \multicolumn{1}{r|}{0.71}        & \multicolumn{1}{r|}{0.69} & \multicolumn{1}{r|}{0.73}           & 0.71     & \multicolumn{1}{r|}{0.75}        & \multicolumn{1}{r|}{0.80} & \multicolumn{1}{r|}{0.77}           & 0.79     \\ \hline
% Likelihood  & \multicolumn{1}{r|}{0.69}        & \multicolumn{1}{r|}{0.66} & \multicolumn{1}{r|}{0.68}           & 0.66     & \multicolumn{1}{r|}{0.59}        & \multicolumn{1}{r|}{0.65} & \multicolumn{1}{r|}{0.71}           & 0.70     \\ \hline
% Diversity   & \multicolumn{1}{r|}{0.70}        & \multicolumn{1}{r|}{0.67} & \multicolumn{1}{r|}{0.75}           & 0.72     & \multicolumn{1}{r|}{0.75}        & \multicolumn{1}{r|}{0.68} & \multicolumn{1}{r|}{0.63}           & 0.64     \\ \hline
% Repetitions & \multicolumn{1}{r|}{0.69}        & \multicolumn{1}{r|}{0.66} & \multicolumn{1}{r|}{0.70}           & 0.69     & \multicolumn{1}{r|}{0.72}        & \multicolumn{1}{r|}{0.66} & \multicolumn{1}{r|}{0.59}           & 0.62     \\ \hline
% \end{tabular}
    % }
    \caption{Comparison of ROC-AUC for confidence scores across different techniques for correct answer classification on LLaVa predictions. Even though accuracy is affected by the method of choice, \avgwbleu{} is better or comparable for all techniques.}
    \label{tab:answer_eq_techniques}
\end{table}

To address this limitation, we replicated the experiments incorporating three similarity techniques for classifying correct answers, chosen based on their widespread adoption and use in comparable experiments~\cite{risch2021semantic}.
A similarity threshold was hand picked upon inspecting $20$ instances.
The results show variations in accuracy. 
Despite these differences, confidence metrics remained stable across the board, with variations of 8\% at most, as shown in \Cref{tab:answer_eq_techniques}. Further details and experiments can be found in \Cref{app:correct_answers}. 

\paragraph{What errors do models make?}

We manually inspected 272 errors made by the best two models on VizWiz-VQA presented in the previous section, one LLM and one LMM, with the goal of getting insights on their weaknesses. We find the following three most frequent type of errors. First, the LLM hallucinates more than the LMM and hallucinated answers in LLMs often trigger a response. Second, the gold caption of the LLM may not include the answer to the question. Third, the answer is true but not useful for visually impaired people. \Cref{app:errors} illustrates each of them.  

%first the blue medicine, 
%second 199 276
%third the OCRA number o 225 price,

\section{Conclusions}

We studied scoring methods to assess the confidence in the predictions of a VQA system in the VizWiz and UNK-VQA datasets. 
We observed that, as shown previously, sampling based methods might present an advantage over likelihood based estimates in text models, but the picture changes for text-image models.
Open ended question with non-entity answers present additional challenges for sampling methods, so the equivalence among possible sampled answers needs to be incorporated as well.
Our \avgwbleu{} score is able to capture the spread of possible samples in a soft way and accomplishes the best results on the calibration evaluations we measured, combining the advantages of sampling and likelihood-based methods.

% In addition, this work was done using a single model family and model size, on a single dataset. It would be valuable to explore to what extent these findings generalize to scenarios beyond our experiments. It is also worth mentioning that the results might differ for models fine-tuned specifically for this task, but we conjecture that this would increase the gap by making the likelihood estimate overly confident, as shown by ~\citet{openai2023gpt4}.

% Finally, it was also clear upon analysis of model errors that the evaluation of prediction correctness was imperfect, relying only on exact match of strings, which is the official metric used for visual question answering. Model-based answer equivalence functions such as the one defined by~\citet{bulian-etal-2022-tomayto} can be used to create a notion of correctness more aligned with human interpretation in future work.

\section*{Limitations}

This study was conducted using several models over two datasets, which might limit the generalization of its conclusions.
No other VQA datasets to our knowledge account for unanswerable questions in the same rich manner. 
For that reason, more analysis and datasets should be created to further research in this space and we hope our results serve as an initial step in that direction.

The use of manually curated captions may introduce a source of bias. Our analysis of error patterns reveals a substantial advantage for the LLMs due to its direct access to these gold captions. Addressing this limitation, future work could explore the feasibility and implications of working with automatically generated captions, allowing for a more realistic assessment in real-world scenarios. %The transition to automatic captions could enhance the generalizability of findings and better align with the challenges encountered in practical applications.

\section*{Ethical Considerations}

In this paper we explore calibration methods for VQA models. This is an under-researched area which is particularly relevant for the visually impaired community for at least two reasons. First, quantifying the uncertainty in the day-to-day questions from this community can help them decide when to trust an automated VQA and when to ask another person. Second, our findings show that metrics in previous work have unstable performances for different kinds of models when evaluated over questions with high uncertainty such as those from visually impaired people.

The use of rating systems and visual question answering in the context of people with low or no vision carries risks since users of such systems cannot easily verify their results, and erroneous answers can lead to serious harm.
In addition to accidental failures, models such as those studied can be attacked with the use of adversarial examples~\cite{alzantot-etal-2018-generating} by malicious actors with the aim of taking advantage of a vulnerable population.

For these reasons, before offering VQA systems for this population, it is essential to conduct detailed studies of the use cases and their possible failures, with emphasis on risk mitigation and respect of vulnerable populations~\cite{le-ferrand-etal-2022-learning}. 
These studies should be designed collaboratively with members of the target group to minimize biases about how said system can be used.

\section*{Acknowledgments}
We would like to thank Francesco Piccinno, Srini Narayanan, Luciana Ferrer, and the anonymous reviewers for their time, constructive feedback, useful comments and suggestions about this work. This work used computational resources from both Google DeepMind and CCAD – Universidad Nacional de Córdoba\footnote{\url{https://ccad.unc.edu.ar}}, which are part of SNCAD – MinCyT, República Argentina.

\bibliography{bib/journal,bib/custom}

\begin{thebibliography}{41}
\expandafter\ifx\csname natexlab\endcsname\relax\def\natexlab#1{#1}\fi

\bibitem[{Alayrac et~al.(2022)Alayrac, Donahue, Luc, Miech, Barr, Hasson, Lenc, Mensch, Millican, Reynolds, Ring, Rutherford, Cabi, Han, Gong, Samangooei, Monteiro, Menick, Borgeaud, Brock, Nematzadeh, Sharifzadeh, Bi\'{n}kowski, Barreira, Vinyals, Zisserman, and Simonyan}]{flamingo2022}
Jean-Baptiste Alayrac, Jeff Donahue, Pauline Luc, Antoine Miech, Iain Barr, Yana Hasson, Karel Lenc, Arthur Mensch, Katherine Millican, Malcolm Reynolds, Roman Ring, Eliza Rutherford, Serkan Cabi, Tengda Han, Zhitao Gong, Sina Samangooei, Marianne Monteiro, Jacob~L Menick, Sebastian Borgeaud, Andy Brock, Aida Nematzadeh, Sahand Sharifzadeh, Miko\l~aj Bi\'{n}kowski, Ricardo Barreira, Oriol Vinyals, Andrew Zisserman, and Kar\'{e}n Simonyan. 2022.
\newblock \href {https://proceedings.neurips.cc/paper_files/paper/2022/file/960a172bc7fbf0177ccccbb411a7d800-Paper-Conference.pdf} {Flamingo: a visual language model for few-shot learning}.
\newblock In \emph{Advances in Neural Information Processing Systems}, volume~35, pages 23716--23736. Curran Associates, Inc.

\bibitem[{Almazrouei et~al.(2023)Almazrouei, Alobeidli, Alshamsi, Cappelli, Cojocaru, Debbah, Étienne Goffinet, Hesslow, Launay, Malartic, Mazzotta, Noune, Pannier, and Penedo}]{almazrouei2023falcon}
Ebtesam Almazrouei, Hamza Alobeidli, Abdulaziz Alshamsi, Alessandro Cappelli, Ruxandra Cojocaru, Mérouane Debbah, Étienne Goffinet, Daniel Hesslow, Julien Launay, Quentin Malartic, Daniele Mazzotta, Badreddine Noune, Baptiste Pannier, and Guilherme Penedo. 2023.
\newblock \href {http://arxiv.org/abs/2311.16867} {The falcon series of open language models}.

\bibitem[{Alzantot et~al.(2018)Alzantot, Sharma, Elgohary, Ho, Srivastava, and Chang}]{alzantot-etal-2018-generating}
Moustafa Alzantot, Yash Sharma, Ahmed Elgohary, Bo-Jhang Ho, Mani Srivastava, and Kai-Wei Chang. 2018.
\newblock \href {https://doi.org/10.18653/v1/D18-1316} {Generating natural language adversarial examples}.
\newblock In \emph{Proceedings of Empirical Methods in Natural Language Processing}, pages 2890--2896. Association for Computational Linguistics.

\bibitem[{Anil et~al.(2023)Anil, Dai, Firat, Johnson, Lepikhin, Passos, Shakeri, Taropa, Bailey, Chen et~al.}]{anil2023palm}
Rohan Anil, Andrew~M Dai, Orhan Firat, Melvin Johnson, Dmitry Lepikhin, Alexandre Passos, Siamak Shakeri, Emanuel Taropa, Paige Bailey, Zhifeng Chen, et~al. 2023.
\newblock \href {http://arxiv.org/abs/2305.10403} {{PaLM} 2 technical report}.

\bibitem[{Antol et~al.(2015)Antol, Agrawal, Lu, Mitchell, Batra, Zitnick, and Parikh}]{VQA}
Stanislaw Antol, Aishwarya Agrawal, Jiasen Lu, Margaret Mitchell, Dhruv Batra, C.~Lawrence Zitnick, and Devi Parikh. 2015.
\newblock \href {https://openaccess.thecvf.com/content_iccv_2015/html/Antol_VQA_Visual_Question_ICCV_2015_paper.html} {{VQA}: {V}isual {Q}uestion {A}nswering}.
\newblock In \emph{International Conference on Computer Vision (ICCV)}.

\bibitem[{Bulian et~al.(2022)Bulian, Buck, Gajewski, B{\"o}rschinger, and Schuster}]{bulian-etal-2022-tomayto}
Jannis Bulian, Christian Buck, Wojciech Gajewski, Benjamin B{\"o}rschinger, and Tal Schuster. 2022.
\newblock \href {https://doi.org/10.18653/v1/2022.emnlp-main.20} {Tomayto, tomahto. beyond token-level answer equivalence for question answering evaluation}.
\newblock In \emph{Proceedings of Empirical Methods in Natural Language Processing}. Association for Computational Linguistics.

\bibitem[{Chen et~al.(2022)Chen, Anjum, and Gurari}]{ChenAG22}
Chongyan Chen, Samreen Anjum, and Danna Gurari. 2022.
\newblock \href {https://doi.org/10.1109/CVPR52688.2022.01851} {Grounding answers for visual questions asked by visually impaired people}.
\newblock In \emph{{IEEE/CVF} Conference on Computer Vision and Pattern Recognition, {CVPR} 2022, New Orleans, LA, USA, June 18-24, 2022}, pages 19076--19085. {IEEE}.

\bibitem[{Chen et~al.(2023{\natexlab{a}})Chen, Anjum, and Gurari}]{Chen_2023_ICCV}
Chongyan Chen, Samreen Anjum, and Danna Gurari. 2023{\natexlab{a}}.
\newblock \href {https://doi.ieeecomputersociety.org/10.1109/ICCV51070.2023.01405} {{VQA} therapy: Exploring answer differences by visually grounding answers}.
\newblock In \emph{Proceedings of the IEEE/CVF International Conference on Computer Vision}, pages 15315--15325.

\bibitem[{Chen et~al.(2023{\natexlab{b}})Chen, Djolonga, Padlewski, Mustafa, Changpinyo, Wu, Ruiz, Goodman, Wang, Tay, Shakeri, Dehghani, Salz, Lucic, Tschannen, Nagrani, Hu, Joshi, Pang, Montgomery, Pietrzyk, Ritter, Piergiovanni, Minderer, Pavetic, Waters, Li, Alabdulmohsin, Beyer, Amelot, Lee, Steiner, Li, Keysers, Arnab, Xu, Rong, Kolesnikov, Seyedhosseini, Angelova, Zhai, Houlsby, and Soricut}]{chen2023palix}
Xi~Chen, Josip Djolonga, Piotr Padlewski, Basil Mustafa, Soravit Changpinyo, Jialin Wu, Carlos~Riquelme Ruiz, Sebastian Goodman, Xiao Wang, Yi~Tay, Siamak Shakeri, Mostafa Dehghani, Daniel Salz, Mario Lucic, Michael Tschannen, Arsha Nagrani, Hexiang Hu, Mandar Joshi, Bo~Pang, Ceslee Montgomery, Paulina Pietrzyk, Marvin Ritter, AJ~Piergiovanni, Matthias Minderer, Filip Pavetic, Austin Waters, Gang Li, Ibrahim Alabdulmohsin, Lucas Beyer, Julien Amelot, Kenton Lee, Andreas~Peter Steiner, Yang Li, Daniel Keysers, Anurag Arnab, Yuanzhong Xu, Keran Rong, Alexander Kolesnikov, Mojtaba Seyedhosseini, Anelia Angelova, Xiaohua Zhai, Neil Houlsby, and Radu Soricut. 2023{\natexlab{b}}.
\newblock \href {http://arxiv.org/abs/2305.18565} {Pali-x: On scaling up a multilingual vision and language model}.

\bibitem[{Chen et~al.(2023{\natexlab{c}})Chen, Wang, Beyer, Kolesnikov, Wu, Voigtlaender, Mustafa, Goodman, Alabdulmohsin, Padlewski, Salz, Xiong, Vlasic, Pavetic, Rong, Yu, Keysers, Zhai, and Soricut}]{chen2023pali3}
Xi~Chen, Xiao Wang, Lucas Beyer, Alexander Kolesnikov, Jialin Wu, Paul Voigtlaender, Basil Mustafa, Sebastian Goodman, Ibrahim Alabdulmohsin, Piotr Padlewski, Daniel Salz, Xi~Xiong, Daniel Vlasic, Filip Pavetic, Keran Rong, Tianli Yu, Daniel Keysers, Xiaohua Zhai, and Radu Soricut. 2023{\natexlab{c}}.
\newblock \href {http://arxiv.org/abs/2310.09199} {{PaLI-3} vision language models: Smaller, faster, stronger}.

\bibitem[{Chung et~al.(2022)Chung, Hou, Longpre, Zoph, Tay, Fedus, Li, Wang, Dehghani, Brahma et~al.}]{chung2022scaling}
Hyung~Won Chung, Le~Hou, Shayne Longpre, Barret Zoph, Yi~Tay, William Fedus, Eric Li, Xuezhi Wang, Mostafa Dehghani, Siddhartha Brahma, et~al. 2022.
\newblock \href {http://arxiv.org/abs/2210.11416} {Scaling instruction-finetuned language models}.

\bibitem[{Cole et~al.(2023{\natexlab{a}})Cole, Zhang, Gillick, Eisenschlos, Dhingra, and Eisenstein}]{flexqa}
Jeremy~R. Cole, Michael J.~Q. Zhang, Daniel Gillick, Julian~Martin Eisenschlos, Bhuwan Dhingra, and Jacob Eisenstein. 2023{\natexlab{a}}.
\newblock \href {https://openreview.net/forum?id=x2W2dKdNI8} {Selectively answering ambiguous questions}.
\newblock In \emph{Proceedings of Empirical Methods in Natural Language Processing}.

\bibitem[{Cole et~al.(2023{\natexlab{b}})Cole, Zhang, Gillick, Eisenschlos, Dhingra, and Eisenstein}]{cole2023selectively}
Jeremy~R. Cole, Michael J.~Q. Zhang, Daniel Gillick, Julian~Martin Eisenschlos, Bhuwan Dhingra, and Jacob Eisenstein. 2023{\natexlab{b}}.
\newblock \href {http://arxiv.org/abs/2305.14613} {Selectively answering ambiguous questions}.

\bibitem[{Collier et~al.(2021)Collier, Mustafa, Kokiopoulou, Jenatton, and Berent}]{collier2021correlated}
Mark Collier, Basil Mustafa, Efi Kokiopoulou, Rodolphe Jenatton, and Jesse Berent. 2021.
\newblock \href {https://doi.org/10.1109/CVPR46437.2021.00160} {Correlated input-dependent label noise in large-scale image classification}.
\newblock In \emph{2021 IEEE/CVF Conference on Computer Vision and Pattern Recognition (CVPR)}, pages 1551--1560.

\bibitem[{Dancette et~al.(2023)Dancette, Whitehead, Maheshwary, Vedantam, Scherer, Chen, Cord, and Rohrbach}]{dancette2023improving}
Corentin Dancette, Spencer Whitehead, Rishabh Maheshwary, Ramakrishna Vedantam, Stefan Scherer, Xinlei Chen, Matthieu Cord, and Marcus Rohrbach. 2023.
\newblock \href {https://doi.org/10.1109/CVPR52729.2023.02303} {Improving selective visual question answering by learning from your peers}.
\newblock In \emph{Computer Vision and Pattern Recognition}, pages 24049--24059, Los Alamitos, CA, USA. IEEE Computer Society.

\bibitem[{Desai and Durrett(2020)}]{desai-durrett-2020-calibration}
Shrey Desai and Greg Durrett. 2020.
\newblock \href {https://doi.org/10.18653/v1/2020.emnlp-main.21} {Calibration of pre-trained transformers}.
\newblock In \emph{Proceedings of Empirical Methods in Natural Language Processing}, pages 295--302. Association for Computational Linguistics.

\bibitem[{Goyal et~al.(2017)Goyal, Khot, Summers{-}Stay, Batra, and Parikh}]{balanced_vqa_v2}
Yash Goyal, Tejas Khot, Douglas Summers{-}Stay, Dhruv Batra, and Devi Parikh. 2017.
\newblock \href {https://openaccess.thecvf.com/content_cvpr_2017/papers/Goyal_Making_the_v_CVPR_2017_paper.pdf} {Making the {V} in {VQA} matter: Elevating the role of image understanding in {V}isual {Q}uestion {A}nswering}.
\newblock In \emph{Computer Vision and Pattern Recognition}.

\bibitem[{Guo et~al.(2017)Guo, Pleiss, Sun, and Weinberger}]{guo2017calibration}
Chuan Guo, Geoff Pleiss, Yu~Sun, and Kilian~Q Weinberger. 2017.
\newblock \href {https://proceedings.mlr.press/v70/guo17a/guo17a.pdf} {On calibration of modern neural networks}.
\newblock In \emph{International conference on machine learning}, pages 1321--1330. PMLR.

\bibitem[{Guo et~al.(2023)Guo, Jiao, Shen, Nie, and Kankanhalli}]{guo2023unkvqa}
Yanyang Guo, Fangkai Jiao, Zhiqi Shen, Liqiang Nie, and Mohan Kankanhalli. 2023.
\newblock \href {http://arxiv.org/abs/2310.10942} {Unk-vqa: A dataset and a probe into multi-modal large models' abstention ability}.

\bibitem[{Gurari et~al.(2018)Gurari, Li, Stangl, Guo, Lin, Grauman, Luo, and Bigham}]{gurari2018vizwiz}
Danna Gurari, Qing Li, Abigale~J Stangl, Anhong Guo, Chi Lin, Kristen Grauman, Jiebo Luo, and Jeffrey~P Bigham. 2018.
\newblock \href {https://openaccess.thecvf.com/content_cvpr_2018/papers/Gurari_VizWiz_Grand_Challenge_CVPR_2018_paper} {Vizwiz grand challenge: Answering visual questions from blind people}.
\newblock \emph{Computer Vision and Pattern Recognition}.

\bibitem[{Gurari et~al.(2020)Gurari, Zhao, Zhang, and Bhattacharya}]{vizwizcaps2020}
Danna Gurari, Yinan Zhao, Meng Zhang, and Nilavra Bhattacharya. 2020.
\newblock \href {https://link.springer.com/chapter/10.1007/978-3-030-58520-4_25} {Captioning images taken by people who are blind}.
\newblock In \emph{Computer Vision - ECCV 2020}, pages 417--434, Cham. Springer International.

\bibitem[{Harnad(1990)}]{harnad:1990}
Stevan Harnad. 1990.
\newblock The symbol grounding problem.
\newblock \emph{Physica D: Nonlinear Phenomena}, 42(1):335 -- 346.

\bibitem[{Jiang et~al.(2021)Jiang, Araki, Ding, and Neubig}]{jiang-etal-2021-how}
Zhengbao Jiang, Jun Araki, Haibo Ding, and Graham Neubig. 2021.
\newblock \href {https://doi.org/10.1162/tacl_a_00407} {{How Can We Know When Language Models Know? On the Calibration of Language Models for Question Answering}}.
\newblock \emph{Transactions of the Association for Computational Linguistics}, 9:962--977.

\bibitem[{Kamath et~al.(2020)Kamath, Jia, and Liang}]{kamath-etal-2020-selective}
Amita Kamath, Robin Jia, and Percy Liang. 2020.
\newblock \href {https://doi.org/10.18653/v1/2020.acl-main.503} {Selective question answering under domain shift}.
\newblock In \emph{Proceedings of the Association for Computational Linguistics}, pages 5684--5696, Online. Association for Computational Linguistics.

\bibitem[{Le~Ferrand et~al.(2022)Le~Ferrand, Bird, and Besacier}]{le-ferrand-etal-2022-learning}
Eric Le~Ferrand, Steven Bird, and Laurent Besacier. 2022.
\newblock \href {https://doi.org/10.18653/v1/2022.acl-long.342} {Learning from failure: Data capture in an {A}ustralian aboriginal community}.
\newblock In \emph{Proceedings of the 60th Annual Meeting of the Association for Computational Linguistics}, pages 4988--4998. Association for Computational Linguistics.

\bibitem[{Li et~al.(2023)Li, Li, Savarese, and Hoi}]{li2023blip2}
Junnan Li, Dongxu Li, Silvio Savarese, and Steven Hoi. 2023.
\newblock \href {http://arxiv.org/abs/2301.12597} {Blip-2: Bootstrapping language-image pre-training with frozen image encoders and large language models}.

\bibitem[{Lin(2004)}]{lin-2004-rouge}
Chin-Yew Lin. 2004.
\newblock \href {https://aclanthology.org/W04-1013} {{ROUGE}: A package for automatic evaluation of summaries}.
\newblock In \emph{Text Summarization Branches Out}, pages 74--81, Barcelona, Spain. Association for Computational Linguistics.

\bibitem[{Liu et~al.(2023)Liu, Li, Li, and Lee}]{llava2023}
Haotian Liu, Chunyuan Li, Yuheng Li, and Yong~Jae Lee. 2023.
\newblock \href {http://arxiv.org/abs/2310.03744} {Improved baselines with visual instruction tuning}.
\newblock In \emph{Proceedings of Advances in Neural Information Processing Systems}.

\bibitem[{Maynez et~al.(2023)Maynez, Agrawal, and Gehrmann}]{maynez-etal-2023-benchmarking}
Joshua Maynez, Priyanka Agrawal, and Sebastian Gehrmann. 2023.
\newblock \href {https://doi.org/10.18653/v1/2023.acl-long.511} {Benchmarking large language model capabilities for conditional generation}.
\newblock In \emph{Proceedings of the Association for Computational Linguistics}, pages 9194--9213, Toronto, Canada. Association for Computational Linguistics.

\bibitem[{Papineni et~al.(2002)Papineni, Roukos, Ward, and Zhu}]{papineni-etal-2002-bleu}
Kishore Papineni, Salim Roukos, Todd Ward, and Wei-Jing Zhu. 2002.
\newblock \href {https://doi.org/10.3115/1073083.1073135} {{BLEU}: a method for automatic evaluation of machine translation}.
\newblock In \emph{Proceedings of the Association for Computational Linguistics}, pages 311--318. Association for Computational Linguistics.

\bibitem[{Rajpurkar et~al.(2016)Rajpurkar, Zhang, Lopyrev, and Liang}]{rajpurkar-etal-2016-squad}
Pranav Rajpurkar, Jian Zhang, Konstantin Lopyrev, and Percy Liang. 2016.
\newblock \href {https://doi.org/10.18653/v1/D16-1264} {{SQ}u{AD}: 100,000+ questions for machine comprehension of text}.
\newblock In \emph{Proceedings of Empirical Methods in Natural Language Processing}, pages 2383--2392, Austin, Texas. Association for Computational Linguistics.

\bibitem[{Rambow and Walker(1994)}]{rambow-walker-1994-role}
Owen Rambow and Marilyn Walker. 1994.
\newblock \href {https://aclanthology.org/W94-0320} {The role of cognitive modeling in communicative intentions}.
\newblock In \emph{Proceedings of the Seventh International Workshop on Natural Language Generation}.

\bibitem[{Risch et~al.(2021)Risch, M{\"o}ller, Gutsch, and Pietsch}]{risch2021semantic}
Julian Risch, Timo M{\"o}ller, Julian Gutsch, and Malte Pietsch. 2021.
\newblock \href {https://doi.org/10.18653/v1/2021.mrqa-1.15} {Semantic answer similarity for evaluating question answering models}.
\newblock In \emph{Proceedings of the 3rd Workshop on Machine Reading for Question Answering}, pages 149--157, Punta Cana, Dominican Republic. Association for Computational Linguistics.

\bibitem[{Sellam et~al.(2020)Sellam, Das, and Parikh}]{sellam-etal-2020-bleurt}
Thibault Sellam, Dipanjan Das, and Ankur Parikh. 2020.
\newblock \href {https://doi.org/10.18653/v1/2020.acl-main.704} {{BLEURT}: Learning robust metrics for text generation}.
\newblock In \emph{Proceedings of the Association for Computational Linguistics}, pages 7881--7892, Online. Association for Computational Linguistics.

\bibitem[{Shazeer and Stern(2018)}]{pmlr-v80-shazeer18a}
Noam Shazeer and Mitchell Stern. 2018.
\newblock \href {https://proceedings.mlr.press/v80/shazeer18a.html} {Adafactor: Adaptive learning rates with sublinear memory cost}.
\newblock In \emph{Proceedings of the 35th International Conference on Machine Learning}, pages 4596--4604. PMLR.

\bibitem[{Wan et~al.(2023)Wan, Sun, Nakhost, Dai, Eisenschlos, Arik, and Pfister}]{wan2023universal}
Xingchen Wan, Ruoxi Sun, Hootan Nakhost, Hanjun Dai, Julian~Martin Eisenschlos, Sercan~O. Arik, and Tomas Pfister. 2023.
\newblock \href {https://arxiv.org/abs/2305.14926} {Universal self-adaptive prompting}.
\newblock In \emph{Proceedings of the 2023 Conference on Empirical Methods in Natural Language Processing}.

\bibitem[{Wang et~al.(2020)Wang, Wei, Dong, Bao, Yang, and Zhou}]{wang2020minilm}
Wenhui Wang, Furu Wei, Li~Dong, Hangbo Bao, Nan Yang, and Ming Zhou. 2020.
\newblock \href {https://proceedings.neurips.cc/paper/2020/hash/3f5ee243547dee91fbd053c1c4a845aa-Abstract.html} {{MiniLM}: Deep self-attention distillation for task-agnostic compression of pre-trained transformers}.
\newblock In \emph{Proceedings of the 34th International Conference on Neural Information Processing Systems}. Curran Associates Inc.

\bibitem[{Wei et~al.(2023)Wei, Yuan, Yang, Shen, Li, Wang, and Chen}]{wei-etal-2023-tackling}
Yiwei Wei, Shaozu Yuan, Ruosong Yang, Lei Shen, Zhangmeizhi Li, Longbiao Wang, and Meng Chen. 2023.
\newblock \href {https://doi.org/10.18653/v1/2023.acl-long.287} {Tackling modality heterogeneity with multi-view calibration network for multimodal sentiment detection}.
\newblock In \emph{Proceedings of the 61st Annual Meeting of the Association for Computational Linguistics}. Association for Computational Linguistics.

\bibitem[{Whitehead et~al.(2022)Whitehead, Petryk, Shakib, Gonzalez, Darrell, Rohrbach, and Rohrbach}]{reliable-vqa}
Spencer Whitehead, Suzanne Petryk, Vedaad Shakib, Joseph Gonzalez, Trevor Darrell, Anna Rohrbach, and Marcus Rohrbach. 2022.
\newblock \href {https://doi.org/10.1007/978-3-031-20059-5_9} {Reliable visual question answering: Abstain rather than answer incorrectly}.
\newblock In \emph{Computer Vision – ECCV 2022}, page 148–166, Berlin, Heidelberg. Springer-Verlag.

\bibitem[{Wilcox(1973)}]{wilcox-1973-iqv}
Allen~R. Wilcox. 1973.
\newblock \href {http://www.jstor.org/stable/446831} {Indices of qualitative variation and political measurement}.
\newblock \emph{The Western Political Quarterly}, 26(2):325--343.

\bibitem[{Zhang et~al.(2020)Zhang, Liao, and Bellamy}]{zhang-etal-2020-effect}
Yunfeng Zhang, Q.~Vera Liao, and Rachel K.~E. Bellamy. 2020.
\newblock \href {https://doi.org/10.1145/3351095.3372852} {Effect of confidence and explanation on accuracy and trust calibration in ai-assisted decision making}.
\newblock In \emph{Proceedings of the 2020 Conference on Fairness, Accountability, and Transparency}, FAT* '20, page 295–305, New York, NY, USA. Association for Computing Machinery.

\end{thebibliography}
\bibliographystyle{acl_natbib}

\clearpage

\appendix

\section{Few-shot Experiments}\label{app:exp}

We show in \Cref{fig_app:llm_prompt} and \Cref{fig_app:lmm_prompt} the $4$-shot prompts used for LLMs and LMMs respectively, with representative examples chosen from the training set.
The experiments were conducted in Tesla T4 GPUs with 16GB of VRAM and it took less than 4 hours for each model we executed.
The results from BLIP-2~\cite{li2023blip2} are shown in \Cref{tab:more_results}.
We used the official models checkpoints released in \url{https://hf.co}.
For PaLM2, we used the publicly available API\footnote{\url{https://cloud.google.com/vertex-ai}} and the Flamingo predictions on the dataset were shared by the authors.

\begin{figure}[ht]
    \centering
    \begin{mdframed}
\begin{lstlisting}[basicstyle=\linespread{0.8}\ttfamily\tiny,breaklines=true,postbreak=\mbox{\textcolor{red}{$\hookrightarrow$}\space}]
Read the descriptions of the following images and answer the questions using a single word or phrase. When the provided information is insufficient, respond with 'unanswerable'. 

Descriptions:
- close up of a computer monitor that is powered on.
- A monitor has a message displayed on it.
- Pictured here is a screenshot that shows an error message from an app.
- Computer screen displaying an error saying the display driver is not supported by Zoom Text.
- a screenshot of someone's monitor that is having issues

Question: What does the arrow say?
Answer: unanswerable

Descriptions:
- a white paper showing an image of black and brown dog
- A library book with pictures of two dogs on the cover on a wooden table.
- A book with a black and a tan dog walking down a snowy street.
- The book cover shows two dogs in the snow
- A book cover title Dog Years with an image of a black and brown dog walking up the street, on the left side it has a due date sticker from a library

Question: What is the title of this book?
Answer: dog years

Descriptions:
- Quality issues are too severe to recognize visual content.
- A white object with elastic sides sitting on a wooden surface.
- A baby diaper keep in the table shown by the image.
- A opened up Diaper laying on a wooden surface.
- Clean maxi Pad over a dark wooden surface.

Question: What color is this
Answer: white

Descriptions:
- image is blur and hard to find what it is
- A woman in pink pajamas standing next to a white refrigerator in a kitchen.
- The back of a woman in pink standing at her counter and next to her white refrigerator
- A photo of a person wearing a pink shirt and pink flower pants in a kitchen next to a refrigerator.
- A person in red shirt and pink pajama pants stands in a kitchen near cabinets and counters.

Question: Is this a woman?
Answer: yes

Descriptions:
{descriptions}

Question: {question}
Answer:
\end{lstlisting}
    \end{mdframed}
    \caption{LLM 4-shot prompting for VizWiz-VQA using captions from ~\citet{vizwizcaps2020}.}
    \label{fig_app:llm_prompt}
\end{figure}

\begin{figure}[ht!]
    \centering
    \begin{mdframed}
\begin{lstlisting}[escapeinside=||,basicstyle=\linespread{0.8}\ttfamily\tiny,breaklines=true,postbreak=\mbox{\textcolor{red}{$\hookrightarrow$}\space}]
Answer the question about the image using a single word or phrase. When the provided information is insufficient, respond with 'unanswerable'. 

|\includegraphics[width=0.5\textwidth]{figures/VizWiz_val_00000004.jpg}|

Question: What does the arrow say?
Answer: unanswerable

|\includegraphics[width=0.5\textwidth]{figures/VizWiz_val_00000036.jpg}|


Question: What is the title of this book?
Answer: dog years

|\includegraphics[width=0.5\textwidth]{figures/VizWiz_val_00000011.jpg}|


Question: What color is this
Answer: white

|\includegraphics[width=0.5\textwidth]{figures/VizWiz_val_00000028.jpg}|


Question: Is this a woman?
Answer: yes

[IMAGE]

Question: {question}
Answer:
\end{lstlisting}
    \end{mdframed}
    \caption{LMM 4-shot prompting for VizWiz-VQA.}
    \label{fig_app:lmm_prompt}
\end{figure}

\section{Fine-tuning experiments}\label{app:finetuning}
The Flan-T5 base model was trained using 3 NVIDIA GTX 1080Ti GPUs (GP102, 11 GiB GDDR5) connected via PCIe 3.0 16x, with 32-bit floating-point precision. The fine-tuning process extended over 4k global steps until convergence was achieved, employing a batch size of 16. The learning rate schedule uses a linear warmup of 800 steps to 1e-5, followed by cosine decay to 0. Model optimization uses Adafactor \cite{pmlr-v80-shazeer18a}. A sequence of 256 tokens was used for input encoding, while the output decoding utilized an 8-token sequence.
The predictions for a fine-tuned PaLI-3 were shared by the authors and we refer to the paper for the training configuration.
We show the results of the experiments in \Cref{tab:fine_tuned}.
\begin{table}[ht]
    \centering
    \footnotesize
    % \resizebox{1.0\textwidth}{!}{%
    \begin{tabular}{lrr}
Model & Acc & Trig \\
\hline
\hline
LLaVA (13B) & 61.2 & 34.0 \\
PaLM-2 (Bison) & 33.6 & 78.8 \\
\hline
PaLI-3 (5B) & 66.7 & 48.1 \\
Flan-T5 (Base) & 74.2 & 55.3 \\
\end{tabular}
    % }
    \caption{Comparison of accuracy and triggering rate for in-context (top) vs fine-tuned (bottom) models. Few-shot LLMs abstain less than the other models.}
    \label{tab:fine_tuned}
\end{table}

\section{Classifying correct answers}\label{app:correct_answers}

As discussed in \Cref{sec:accuracy_methods}, the experiments showcased in this paper used exact-match criteria to classify an answer as correct: only if the output generated by the model was exactly one of the accepted answers it counted as correct. In this section we describe the replication of these experiments, employing identical methodologies, introducing four distinct techniques for classifying correct answers. The selection of these techniques was based on their widespread adoption and their utilization in a comparable experiment outlined in \citep{risch2021semantic}. We perform this experiment for the LLaVA 13B and PaLM 2 Bison with similar results.

Each technique is briefly described as follows: Exact Match (EM) returns True only if the model's answer is exactly one of the accepted answers from the dataset. For Cosine Similarity we used \textit{all-MiniLM-L6-v2} \citep{wang2020minilm} as an embedding model due to its widespread use. BEM, as proposed in \citep{bulian-etal-2022-tomayto}, is a fine-tuned BERT model to classify answer equivalence using the SQuAD Dataset \citep{rajpurkar-etal-2016-squad}.

\begin{table}[t]
    \centering
    \footnotesize
    % \resizebox{1.0\textwidth}{!}{%
    \begin{tabular}{lrrrr}
Method & EM & BLEU & Cos Sim & BEM \\
\hline
Accuracy & 79\% & 81\% & 87\% & 87\% \\
\hline
\textsc{\avgwbleu{}} & 28.6 & 35.42 & 50.57 & 50.18 \\
Likelihood & \underline{7.90} & \underline{10.95} & 24.06 & 22.97 \\
Diversity & \textbf{6.80} & \textbf{6.17} & \textbf{13.39} & \textbf{12.30} \\
Repetitions & 9.9 & 11.20 & \underline{20.49} & \underline{19.62} \\
\end{tabular}
    % }
    \caption{Comparison of ECE metric for confidence scores across different techniques for correct answer classification in LLaVa 13B. Accuracy and ECE metric are affected by the method of choice, nonetheless the best performant method (highlighted in bold) remains constant.}
    \label{tab:app_answer_eq_technique_ECE_LLaVa}
\end{table}

In \Cref{tab:app_answer_eq_technique_ECE_LLaVa} we compare the performance of the ECE method throughout the four tested techniques to classify correct answers. We can observe how both the accuracy and the value of ECE change, however the best performant method remains constant. These findings suggest that the results presented in this paper are robust across various answer classification techniques.

\section{Insights from error analysis}\label{app:errors}

\begin{figure}[!ht] 
\vspace*{.3cm}
\begin{center}
\includegraphics[width=7cm]{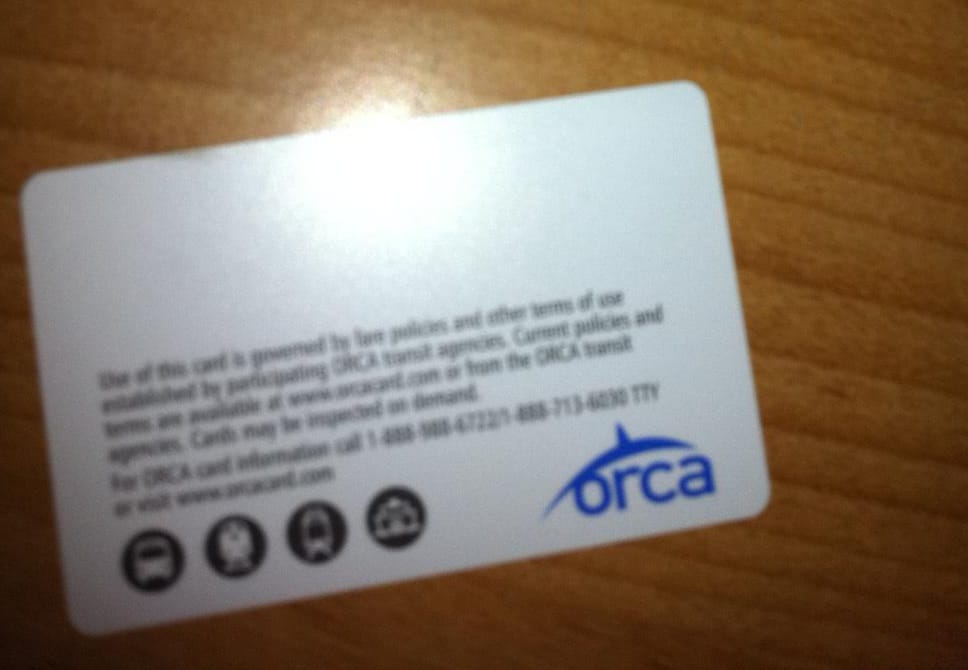}
\caption{Question \emph{``Alright see if you can see the ORCA serial number now.''}. The top 3 answers by PaLM2 Bison with their likelihood are ('31631036', -1.32), ('468563995', -1.41), ('4667926374', -1.41). They are neither in the image nor in the gold captions. The serial number is not in the image. Correctly, LLaVA 13B does not trigger a response in this case considering the question unanswerable.\label{fig:hallucination}}
\end{center}
\end{figure}

In examining 272 errors from the top-performing language models LLaVA 13B which is an LMM and PaLM 2 Bison which is a LLM, we aimed to identify weaknesses specific to the needs of the vulnerable population. Three recurrent errors were observed which accounted for most errors. Firstly, the LLM tends to hallucinate more than the LMM, with these hallucinated responses influencing the final output. For example \Cref{fig:hallucination} illustrates how PaLM 2 Bison hallucinates multiple serial numbers that appear neither in the image nor in the captions. This is less frequent for LLaVA 13B which refrains from answering in most of these cases.

\begin{figure}[ht!] 
\vspace*{.3cm}
\begin{center}
\includegraphics[width=7cm]{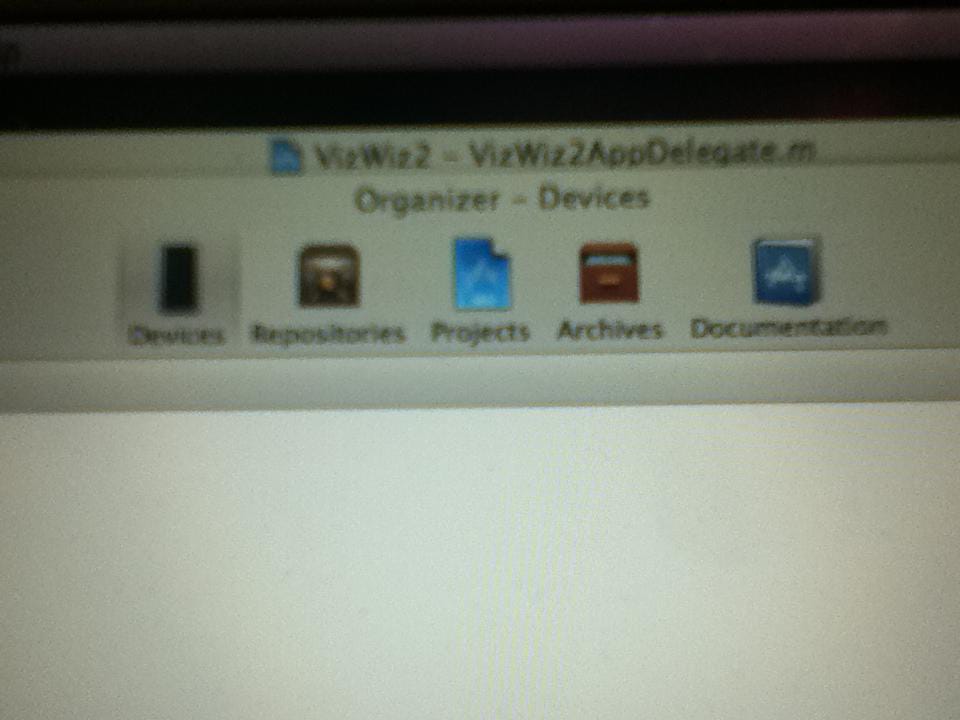}
\caption{Question \emph{``What action is currently selected?''}. The top 3 answers by PaLM2 Bison with their likelihood are ('projects', -0.69), ('archives', -0.76), ('devices', -1.13). The correct answer is devices but this information is not in the gold captions, captions can never be complete. The captions are 'The computer digital monitor screen with some websites open to', 'A quick link bar has appeared on an Apple computer screen.', 'Computer menu bar with options: Devices, Repositories, Projects, Archives, and Documentation.', 'some type of MacBook or laptop device u CNA use', 'A VizWiz app that has an organizer and lists projects and archives.'. \label{fig:missing_info}}
\end{center}
\end{figure}

Secondly, \Cref{fig:missing_info} shows an example of a question that is not possible to answer with the information in the gold captions. This is to be expected a caption cannot include all the information present in an image. This is a limitation of the methodology of using captions instead of images for VQA. 

\begin{figure}[ht!] 
\vspace*{.3cm}
\begin{center}
\includegraphics[width=7cm]{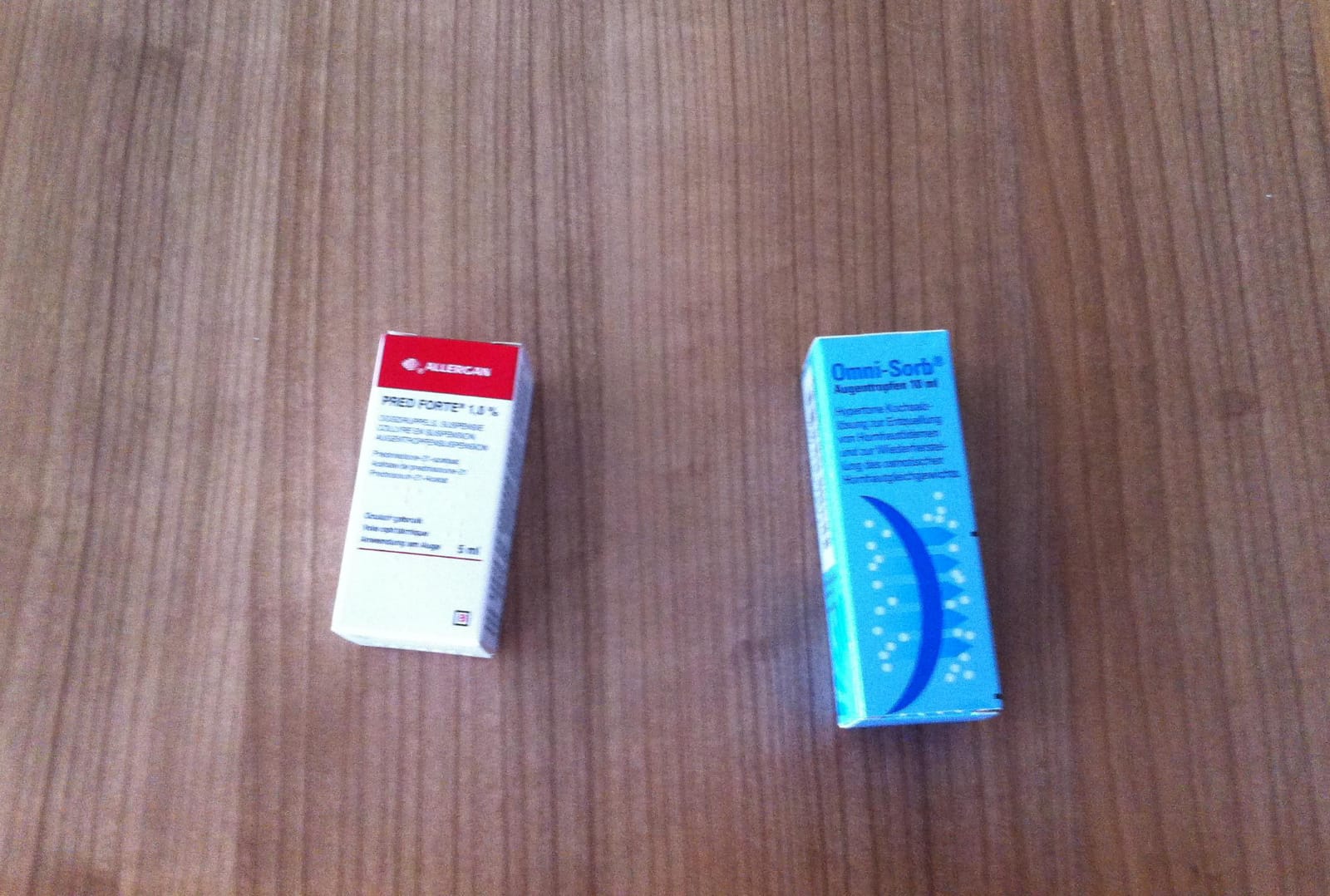}
\caption{Question \emph{``Which one is the blue one?''}.The top 3 answers by PaLM2 Bison with their likelihood are ('blue', -0.47), ('second', -0.63), ('right', -0.82). The first answer is true but useless for a visually impaired person. The top 3 answers by LLaVA 13B Bison are correct. They are ('right', -0.08), ('right', -0.08), ('right', -0.08).  person.\label{fig:useless}}
\end{center}
\end{figure}

Lastly, the third kind of frequent error observed are answers that are factually correct, but are not useful for individuals with visual impairments. A example of this kind of error is shown in \Cref{fig:useless}.

These findings shed light on critical areas for improvement in models designed for this particular user group, showing that well calibrated systems are important in this domain. 

\newpage

\end{document}